\pdfoutput=1

\documentclass[11pt]{article}

\usepackage[final]{acl}

\usepackage{times}
\usepackage{latexsym}

\usepackage[T1]{fontenc}

\usepackage[utf8]{inputenc}

\usepackage{microtype}

\usepackage{inconsolata}

\usepackage{graphicx}

\usepackage{booktabs}
\usepackage{comment}
\usepackage{lipsum}
\usepackage{float}

\newcommand{\new}[1]{\textcolor{black}{#1}}

\title{Experiential Semantic Information and Brain Alignment:\\Are Multimodal Models Better than Language Models?}

\author{Anna Bavaresco, Raquel Fern\'andez \\Institute for Logic, Language and Computation \\ University of Amsterdam \\ \texttt{ \{a.bavaresco, raquel.fernandez\}@uva.nl}}

\begin{document}
\maketitle
\begin{abstract}

    A common assumption in Computational Linguistics is that text representations learnt by multimodal models are richer and more human-like than those by language-only models, as they are grounded in images or audio---similar to how human language is grounded in real-world experiences. However, empirical studies checking whether this is true are largely lacking.
    We address this gap by comparing word representations from contrastive multimodal models vs.~language-only ones in the extent to which they capture experiential information---as defined by an existing norm-based `experiential model'---and align with human fMRI responses. Our results indicate that, surprisingly, language-only models are superior to multimodal ones in both respects. Additionally, they learn more unique brain-relevant semantic information beyond that shared with the experiential model. Overall, our study highlights the need to develop computational models that better integrate the complementary semantic information provided by multimodal data sources.

    \vspace{.11em}
    \hspace{1.25em}\includegraphics[width=1.25em,height=1.25em]{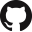}{\hspace{.75em}\parbox{\dimexpr\linewidth-2\fboxsep-2\fboxrule}
    {\url{https:/github.com/dmg-illc/exp-info-models-brain}}}
      
\end{abstract}

\section{Introduction}
\label{sec:introduction}

\new{How to link language representations to the real-world entities they refer to is a long-standing issue within semantics---the `symbol-grounding problem' \cite{harnad1990symbol, bender2020climbing}. With the advent of large language models (LLMs) learning astounding linguistic abilities purely from text, this question has been reframed as the `vector-grounding problem' \cite{mollo2023vector}, gaining new relevance.}
While some researchers think that word meanings should be intended as purely symbolic \cite{fodor1983modularity}, others believe that words have meanings \textit{precisely because} they are linked to specific entities, experiences or notions \cite{barsalou2008grounded}. Supporters of the latter view stress that human language acquisition is situated in a rich multimodal environment, where new words are learnt through interactions with objects and people \cite{vigliocco2014language}. Theories of embodied cognition further highlight the importance of linking words to concrete experience not only for their acquisition but also for their comprehension. Indeed, according to these theories, understanding sentences involves engaging perceptual, motor or emotional simulations of their content \cite[for an overview, see][]{kaschak2024embodied}.  

The idea of obtaining richer semantic representations by learning them from sources other than text, such as images or audio, has inspired a great deal of computational work, from early attempts at concatenating image and text embeddings \cite[e.g.,][]{bruni2014multimodal, kiela2014learning, derby-etal-2018-using, davis-etal-2019-deconstructing} to the most recent large vision-language models \cite[LVLMs, e.g.,][]{li2023blip, wang2024qwen2, liu2024improved, deitke2024molmo, laurenccon2025matters}. Some of these works aimed to obtain language representations aligning more closely with human responses, such as similarity judgments, while others were more oriented towards improving performance on benchmarks or downstream applications. Regardless of the end goal, all these works present multimodality as a \textit{desideratum}, assuming that images provide additional semantic information that cannot be learnt from text alone; however, there is little to no work investigating \textit{which} these semantic aspects are. In this paper, we aim to fill this gap by addressing the following question: \textit{Do multimodal models learn some facets of meaning related to perceptual experiences that language-only models cannot capture?}

Operationalising the `extra-linguistic' information that multimodal models are allegedly learning is a prerequisite for approaching this issue. We did this by relying on a semantic model introduced by \citet{fernandino2022decoding} to capture `experiential information'. 
This cognitive model represents words as $n$-dimensional arrays where each entry corresponds to aggregated human ratings on a pre-defined experiential attribute (e.g., \textit{Vision}, \textit{Motion}, \textit{Harm}).
We focused on a set of nouns and evaluated the alignment between their representations provided by the experiential model and those by comparable unimodal (language-only) and multimodal (vision-language and audio-language) computational models.
This analysis allowed us to uncover if multimodal models indeed reflect more semantic information than language-only models. Next, we checked whether capturing experiential information translates into higher alignment with brain responses recorded with functional magnetic resonance imaging (fMRI) to the same set of nouns. 

Our findings indicate several interesting trends. First, both vision-language and language-only models exhibit significant alignment with the experiential model and brain responses, while the audio-language model displays weak or non-significant correlations. Second, this alignment is more pronounced for language-only models, which appear to capture a great deal of brain-relevant information beyond experiential. Lastly, language-only models remain more brain-aligned than vision-language models even when focusing on a set of more concrete words, although the gap is reduced. Overall, our study shows that current multimodal models learn \textit{less} brain-relevant information---both experiential and beyond---than comparable language-only models, highlighting the need to explore different approaches to construct multimodal word representations. 

\begin{figure*}
    \centering
    \includegraphics[width=\linewidth]{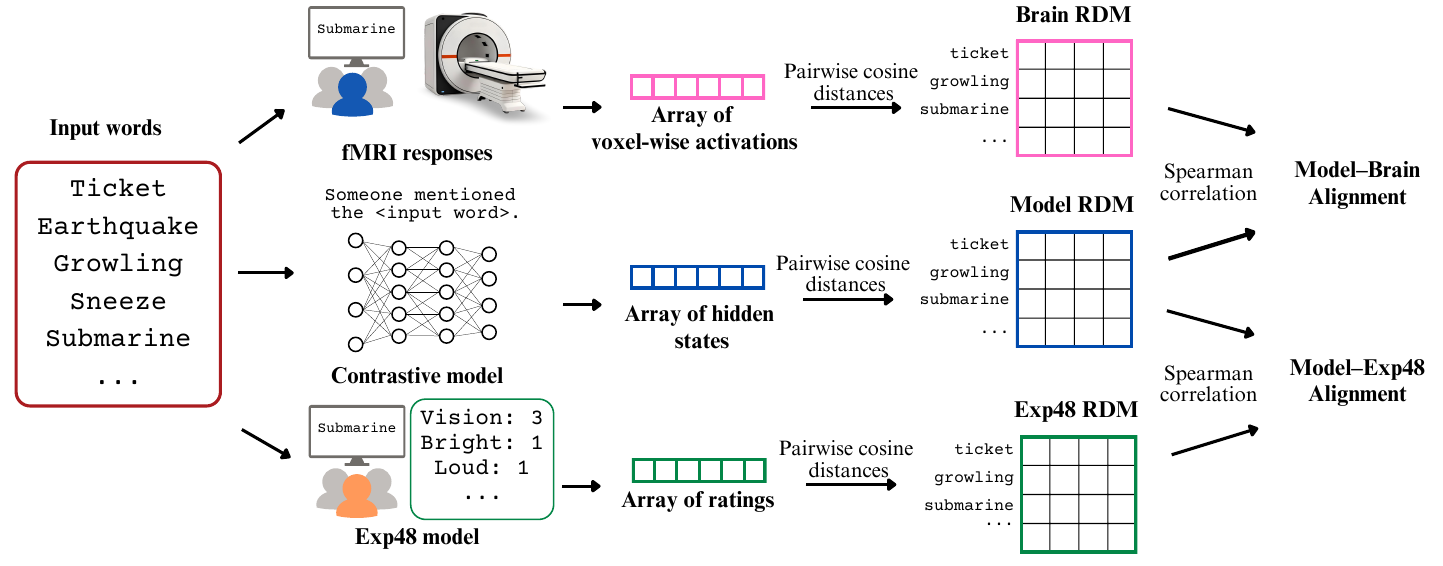}
    \caption{\new{Overview of our experimental setup. Representations for the word stimuli are derived from three different sources: 1) fMRI responses; 2) multimodal and language-only contrastive models; 3) human ratings along the experiential dimensions of the \textsc{Exp48} model. Next, pairwise distances between these word representations are used to populate representational dissimilarity matrices (RDMs). Finally, alignment between representational spaces is computed by correlating the off-diagonal elements of the RDMs.}}
    \label{fig:exp-setup}
\end{figure*}
\section{Background}
\label{sec:background}

\subsection{Embodied cognition}


\textit{Embodied cognition} identifies a suite of theoretical frameworks holding that language is understood by perceptual, emotional, or motor simulations of its content \cite[e.g.,][]{barsalou1999perceptual, glenberg2012action, zwaan2014embodiment, pulvermuller2018neural}. This general principle has received empirical support from multiple studies, both behavioural and neuroscientific. 

For example, a series of works on the Action-sentence Compatibility Effect \cite[ACE,][]{glenberg2002grounding} and its subsequent variants \cite{borreggine2006action, zwaan2006seeing, bub2012dynamics} revealed a significant difference in reaction times---attributed to motor simulations---when participants had to respond to a sentence (e.g., \textit{You passed the note to Art}) with a movement matching (extending their arm) vs.~ non-matching (retreating their arm) that mentioned in the sentence. Similarly, the sentence-picture verification task \cite{stanfield2001effect}, where participants have to respond to a picture that is either compatible (an eagle with its wings outstretched) or incompatible (an eagle with its wings folded) with a sentence (\textit{The eagle is in the sky}), and its variations \cite{connell2007representing, hoeben2017color} have also been widely used to demonstrate the occurrence of perceptual simulation during language comprehension.
In parallel, a line of neuroscientific studies have found evidence that semantic processing may activate motor \cite[among others,][]{hauk2004somatotopic, tettamanti2005listening, aziz2006congruent} and perceptual brain regions \cite{kiefer2008sound, van2012flexibility}.

\subsection{Multimodal models of semantics}



Embodied cognition and related ideas, such as \textit{visual grounding}, have percolated from Cognitive Science to  Computational Linguistics, motivating attempts to build semantic models that learn representations from data sources beyond text. Early efforts in this direction \cite[e.g.,][]{bruni2014multimodal, kiela2014learning, lazaridou2015combining, silberer2012grounded, silberer2014learning} were characterised by 1) a focus on developing human-aligned computational models of meaning and 2) limited computational modelling resources (large datasets of paired image-text inputs did not exist at the time, nor did large transformer-based architectures). 

Recently, multimodal models have become more powerful and found application on a variety of downstream tasks (e.g., image captioning, image retrieval, or visual question answering).
Some seminal works used a contrastive objective to learn aligned image and text representations \cite{radford2021learning, align}, while others---often inspired by BERT's \cite{bert} successes in language modelling---applied its underlying intuitions to the vision-language domain \cite{tan2019lxmert, li2019visualbert, lu2019vilbert, chen2020uniter}. Finally, state-of-the-art large vision-language models  \cite[LVLMs, e.g.,][]{li2023blip, wang2024qwen2, liu2024improved, deitke2024molmo, laurenccon2025matters}, usually combining a large language model (LLM) with an image encoder, can engage in strikingly human-like conversations about images. In contrast to the early attempts at multimodal modelling, these works share 1) a focus on solving, or improving performance on, downstream tasks, and 2) the availability of massive datasets and large models with billions of parameters. 

For our experiments, we aimed to leverage models that are powerful while, at the same time, suitable for drawing cognitively-meaningful comparisons. The need to satisfy both constraints prevented us from evaluating state-of-the-art LVLMs; we elaborate more on our model choices in Section~\ref{sec:comp_models}.\looseness-1 

\subsection{Experiential models of semantics}

Recently, a few approaches motivated by embodied cognition have introduced models of semantics aimed at capturing `experiential information', i.e., aspects of meaning related to how humans ground language in experiences. These experiential models were constructed by asking human annotators to rank words on a set of pre-defined dimensions. For example, \citet{fernandino2022decoding} introduced an experiential model based on 48 dimensions spanning perceptual, emotional, and action-related constructs. In two fMRI studies, they found that the experiential model yields more brain-aligned word representations than taxonomic and distributional models; additionally, it contributes unique semantic information not represented by the other models.

Similarly, \citet{carota2024experientially} experimented with a different experiential model based on 11 dimensions and compared its brain alignment against that of a distributional model. Their study revealed significant correlations with brain responses in more ROIs (regions of interest) for the experiential model than for the distributional model. However, an integrative model combining both displayed significant correlations in an even larger number of ROIs, suggesting that experiential and distributional are complementary aspects of human semantic processing. 

Despite their merits, experiential models are bounded in their accuracy by an \textit{a priori} selection of dimensions and, relying on human annotations, remain expensive to construct. These limitations open the intriguing question of whether experiential information can be captured by computational models learning semantic representations in a data-driven fashion.

\section{Methods}
\label{sec:methods}

\new{A schematic of our experimental pipeline is provided in Figure~\ref{fig:exp-setup}. In the following, we describe in detail the word stimuli, brain responses, computational models and evaluation procedures.}

\subsection{Data and experiential model}
\label{sec:data_and_exp_model}

For our experiments, we used word stimuli, fMRI responses and experiential model from Study 2 by \citet{fernandino2022decoding}.\footnote{These materials have been made publicly available by \citeauthor{fernandino2022decoding} The full list of words and the experiential features can be found at \url{https:/www.pnas.org/doi/10.1073/pnas.2108091119\#supplementary-materials}; fMRI data are available at \url{https:/osf.io/87chb/}.} We describe each below. 

\paragraph{Word stimuli} Word stimuli comprise
320 nouns, half of which refer to \textit{objects} and the other half to \textit{events}. The 160 object nouns include an equal number of words (40) from four categories (\textit{food}, \textit{vehicles}, \textit{animals}, \textit{tools}); likewise, the event nouns span four semantic subcategories (\textit{social event}, \textit{negative event}, \textit{sound}, \textit{communication}).

\paragraph{fMRI responses}
fMRI responses were collected from $36$ participants. While viewing the above-mentioned word stimuli one at a time, they were instructed to rate the frequency with which they experienced their corresponding entities in daily life. \new{Voxel-wise activations (beta maps) for each noun relative to the mean signal across other nouns were estimated using linear regressions \cite[for additional details, see][]{fernandino2022decoding}. Here, we focus on the betas from voxels
within a `semantic network ROI' defined by \citet{binder2009semantic} based on a meta-analysis. Voxel-wise beta coefficients can be arranged in vectors representing the brain response elicited by each noun.}


\paragraph{Experiential model}
The experiential model, hereafter abbreviated as \textsc{Exp48}, represents each word as a set of ratings on 48 pre-defined dimensions capturing different aspects of people's experience with objects/events, e.g., \textit{Vision}, \textit{Hand action} or \textit{Unpleasant}. The ratings were introduced by \citet{binder2016toward} as part of a wider set of experiential salience norms; they range from 0 to 6 and were provided by 1743 unique crowdworkers.

\subsection{Computational models}
\label{sec:comp_models}
 
Our model choices were motivated by the goal to maximise comparability across architectures. More concretely, we selected three models (language-only, vision-language, and audio-language) comparable in terms of fine-tuning objective---the contrastive one---and architecture---they all have a pretrained BERT \cite{bert} as language encoder.\footnote{All three models were released with both BERT-based \cite{bert} and RoBERTa-based \cite{liu2019roberta} implementations. We used the former in all our experiments.} One aspect in which these architectures differ is the amount of training data; however, we believe this issue does not invalidate our results and further discuss it in Section~\ref{sec:disc_concl}.

\paragraph{SimCSE} \cite[\textbf{Sim}ple \textbf{C}ontrastive Learning of \textbf{S}entence \textbf{E}mbeddings,][]{simcse} is a language-only sentence encoder fine-tuned contrastively on 1M sentences randomly sampled from English Wikipedia. Matching pairs for the contrastive objective were created by applying different dropout masks to the same sentence. 

\paragraph{MCSE} \cite[\textbf{M}ultimodal \textbf{C}ontrastive Learning of \textbf{S}entence \textbf{E}mbeddings,][]{zhang-etal-2022-mcse} is a vision-language sentence encoder fine-tuned by jointly optimising a SimCSE objective and a CLIP-like \cite{radford2021learning} objective. The fine-tuning data for the first objective is the same as SimCSE's; as for the CLIP-like objective, where a matching pair was defined by an image and its caption, the fine-tuning data consists of 83K images from MS-COCO \cite{lin2014microsoft} annotated with multiple captions. 


\paragraph{CLAP} \cite[\textbf{C}ontrastive \textbf{L}anguage \textbf{A}udio \textbf{P}retraining,][]{clap} is an audio-language model whose language encoder was initialised with pre-trained BERT weights and fine-tuned on audio-caption pairs with a CLIP-like objective. The fine-tuning data includes $633,526$ audio-text pairs, with audio clips representing human activities, natural sounds, and audio effects. \newline

For reference, we also tested BERT and VisualBERT \cite{li2019visualbert} as its visual counterpart. 

\paragraph{BERT} \cite{bert} is a transformer-based language-only model pretrained with two objectives: masked language modelling and next sentence prediction. Its pretraining data includes the BooksCorpus \cite[800M words,][]{zhu2015aligning} and English Wikipedia (2500 words). As mentioned above, SimCSE, MCSE and CLAP fine-tuned pretrained BERT architectures.

\paragraph{VisualBERT} \cite{li2019visualbert} is a vision-language model consisting of a BERT-based language encoder (initialised with parameters from pretrained BERT) and a pretrained visual feature extractor based on Faster RCNN \citep{ren2015faster}. Its training objectives, which mirror BERT's, were masked language modelling with image input and sentence-image prediction. The vision-language pretraining data comprises MS-COCO 
and VQA 2.0 \cite{goyal2017making}. Note that this is \textit{not} a contrastive model; we included it for reference as it can be considered as a vision-language extension of BERT, but it is not directly comparable with MCSE, SimCSE and CLAP. 

\begin{figure*}[h]
    \centering
    \begin{tabular}{cc}
        \includegraphics[width=0.45\textwidth]{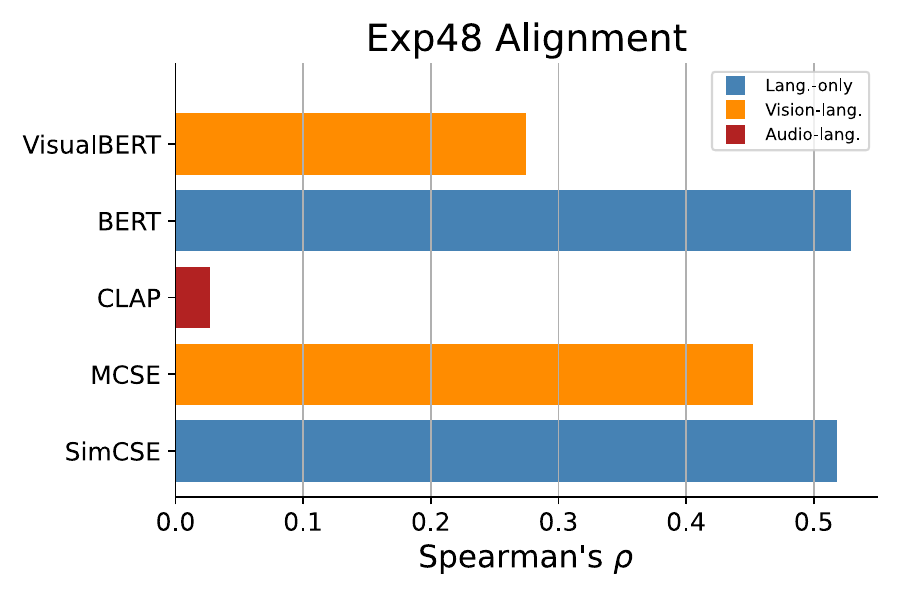} & \includegraphics[width=0.45\textwidth]{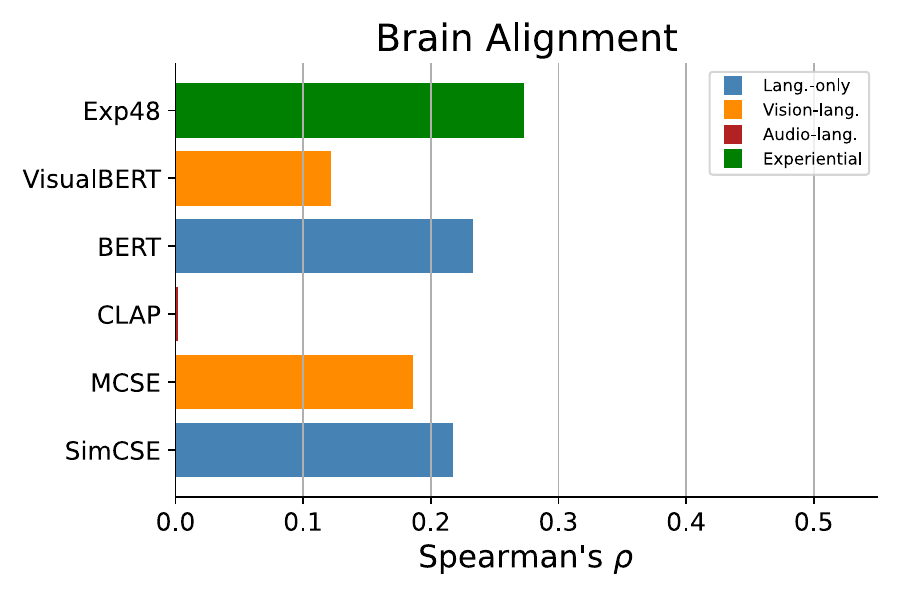} \\

    \end{tabular}
    \caption{Results from representational similarity analysis. On the left, Spearman correlations quantifying the alignment between word representations from \textsc{Exp48} and by computational models. On the right, Spearman correlations indicating the alignment between fMRI responses from human participants and word representations by computational models.}
    \label{fig:correlations}
\end{figure*}

\subsection{Extracting representations}

Given that all the models we considered were trained to learn contextualised representations from sentences, single words may be an out-of-distribution input. Therefore, following an approach similar to \citet{may2019measuring}, we embedded the noun stimuli in a set of generic template sentences (e.g., \texttt{Someone mentioned the <word>}, see Appendix~\ref{app:sentence_templates} for the complete list) when passing them to the models.\footnote{We empirically verified that passing words within templates rather than in isolation yields higher alignment with both the experiential model and brain responses (see Appendix~\ref{app:single_word}).}  
For all templates, we derived word representations from the hidden states of each layer; more specifically, we selected the hidden states corresponding to the tokens of the target word and averaged them across templates. 

\subsection{Alignment evaluation}
\label{sec:alignment_evaluation}

\paragraph{\new{RSA}} To compare model representations against \textsc{Exp48} and brain responses, we used representational similarity analysis \cite[RSA,][]{kriegeskorte2008representational}, which quantifies the alignment between two representational spaces (either by two models or by a model and brain responses) as the correlation between representational dissimilarity matrices (RDMs). In our experiments, RDMs were populated with pairwise cosine distances between model representations or fMRI responses for all the unique word-pairs. fMRI RDMs for individual participants were averaged into one aggregated RDM. The alignment between this fMRI RDM and morel-derived RDMs was calculated as a Spearman correlation.  

\paragraph{\new{Partial correlations}} While RSA allows comparing models' alignment with \textsc{Exp48} or brain responses, it does not reveal whether models explain shared variance or provide independent contributions. 
\citet{fernandino2022decoding} computed partial correlations to check how much brain-relevant information \textsc{Exp48} shared with the other models they considered, i.e., two distributional models \cite[Word2vec and GloVe;][]{word2vec, pennington2014glove} and two taxonomic models (a WordNet-based model and a categorical one). We used the same approach to determine how much brain-relevant information our tested models share with \textsc{Exp48} and with each other. \new{Formally, partial correlations can be defined as follows: Consider the RDM from Model A $y$, the RDM from Model B $x$, and the RDM of the brain responses $z$. The partial correlation of Model A without Model B is $\rho(r_{i},z_{i})$, where $r_{i} = y_{i} - \hat{y}_{i}$ are the residuals from the linear regression with equation $\hat{y}_{i} = a + bx_{i}$.}


\section{Results}
\label{sec:results}

\subsection{\textsc{Exp48}  and brain alignment across models}

We performed RSA to obtain a first measure of model representations' alignment with \textsc{Exp48} and fMRI responses. This analysis was conducted on model representations averaged across the three layers yielding the highest alignment individually; note that these layers may differ when considering alignment to brain responses vs.~\textsc{Exp48} (see Appendix~\ref{app:layer_rsa} for a visualisation of layer-wise alignment). The results from RSA against brain responses and \textsc{Exp48} are displayed in Figure~\ref{fig:correlations}. All Spearman correlations are statistically significant ($p$ < 0.05), except for CLAP's correlation with brain responses ($p$ = 0.70); we additionally verified that all the pairwise differences between correlations are statistically significant.\footnote{Statistical significance was determined by applying a Fisher transformation to the correlation coefficients from each pair of models and calculating the $p$-value associated with the difference between the two $z$-scores. All $p$-values were Bonferroni-corrected with $\alpha$ = 0.05. The same approach for verifying statistical significance was applied to all correlation comparisons throughout the paper.} 

An inspection of correlations against \textsc{Exp48} indicates BERT as the most aligned model  ($\rho$ = 0.53); SimCSE and MCSE also display moderate correlations with \textsc{Exp48} ($\rho$ = 0.52 and $\rho$ = 0.45, respectively). In contrast, CLAP's representations are poorly aligned with \textsc{Exp48}, exhibiting a correlation of just 0.03. A comparison between vision-language models (MCSE and VisualBERT) and their unimodal counterparts (SimCSE and BERT) reveals that the former, surprisingly, reflect less experiential information than the latter.

Regarding alignment with brain responses in the semantic ROI, BERT is again the best model ($\rho$ = 0.23), although it remains less brain-aligned than \textsc{Exp48} ($\rho$ = 0.27). All the other models display positive correlations, with the exception of CLAP, whose correlation is not statistically significant ($\rho$ = 0.00, $p$ = 0.70). Similarly to the \textsc{Exp48}-alignment results, here we found that the language-only models BERT and SimCSE are more brain-aligned than their vision-language extensions VisualBERT and MCSE. We delve deeper into the robustness of this finding in Section~\ref{sec:additional_analyses}. 

\begin{figure*}[]
    \centering
    \begin{tabular}{cc}
        \vtop{\vskip 0pt \hbox{\includegraphics[width=0.425\textwidth]{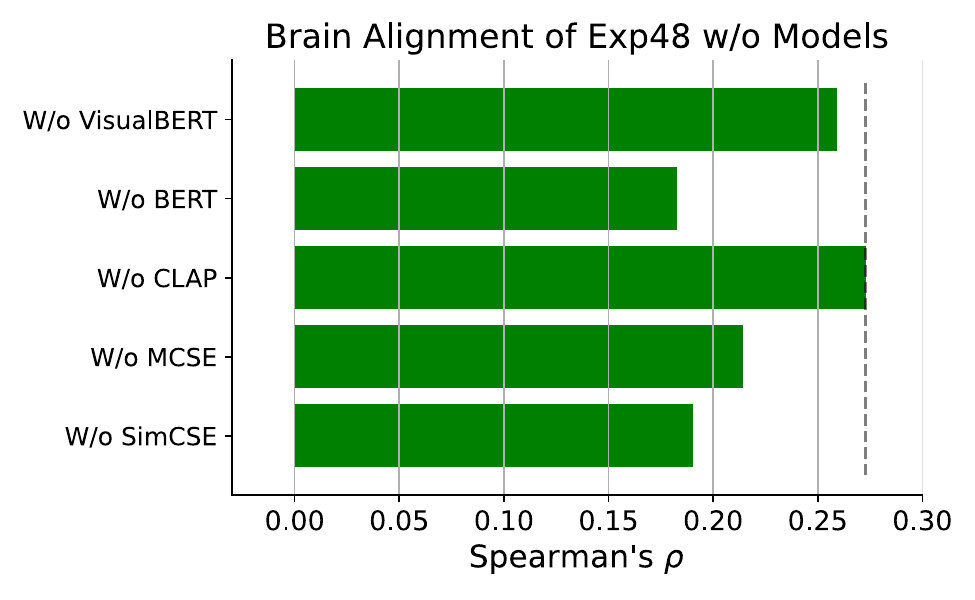}}} & 
        \vtop{\vskip 0pt \hbox{\includegraphics[width=0.4\textwidth]{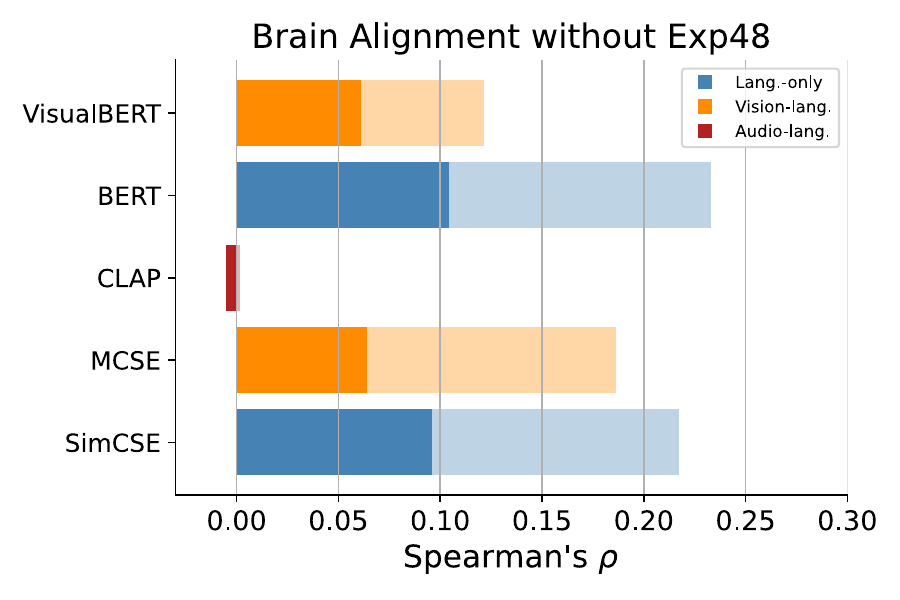}}} \\
    \end{tabular}
    \caption{Results from partial correlation analyses. On the left, Spearman correlations between brain responses and the residuals obtained regressing model RDMs out of the \textsc{Exp48} RDM. The dotted line indicates \textsc{Exp48}'s initial brain alignment without removing any information. On the right, Spearman correlations between brain responses and the residuals obtained regressing the \textsc{Exp48} RDM
    out of model RDMs. The bars in lighter shades indicate models' initial brain alignment.}
    \label{fig:partial_correlations}
\end{figure*}

An interesting trend common across results from both RSAs (against \textsc{Exp48} and fMRI responses) is that representations by SimCSE and MCSE---which have been shown to outperform BERT on semantic text similarity tasks \cite{simcse, zhang-etal-2022-mcse}---are \textit{less} aligned than those by BERT. A potential explanation for this may be that we considered \textit{single-word} representations. Since contrastive fine-tuning, as applied to SimCSE and MCSE, optimises \textit{sentence}-level representations as opposed to \textit{token}-level ones, it could be that some token-level semantic properties initially learnt by BERT got somehow diluted through this process. 

\subsection{Experiential information vs.~unique contribution in models' brain alignment}

Results from the partial correlation analysis are displayed in Figure~\ref{fig:partial_correlations},
whose left-hand panel shows how much \textsc{Exp48} representations align with brain responses without the information they share with each of the other models.
An interesting observation is that the lowest correlations were obtained when regressing out BERT and SimCSE. This provides an interesting complement to the findings from RSA against \textsc{Exp48} representations: RSA shows that BERT and SimCSE share substantial representational information with \textsc{Exp48}, and partial correlations suggest that this information is also brain-relevant. Regarding models' brain alignment without \textsc{Exp48}, displayed in Figure~\ref{fig:partial_correlations}'s right-hand panel, a noteworthy finding is that BERT's and SimCSE's representations are the most brain-aligned even after regressing out \textsc{Exp48}. This suggests that these models learnt some semantic information that is not captured by \textsc{Exp48} but is still reflected in brain responses.

Additionally, for each model we checked which proportion of its initial brain alignment is attributable to unique contribution as opposed to information shared with \textsc{Exp48}. This can be visualised by comparing the dark-shade bars against the light-shade ones in the right-hand panel of Figure~\ref{fig:partial_correlations}. An interesting result revealed by this comparison is that, although MCSE is more brain-aligned than VisualBERT, their unique contribution without \textsc{Exp48} is the same in absolute value ($\rho$ = 0.06); in other terms, 50\% of VisualBERT's brain alignment is due to unique information, while in MCSE it is 32\%. Regarding BERT and SimCSE, the majority of their initial brain alignment is eroded when regressing out \textsc{Exp48}; however, the asymmetry is not substantial, and the unique contribution accounts for more than 40\% of the initial brain alignment in both models. As for CLAP, it exhibits a weak negative correlation that is not statistically significant, confirming that the model does not contribute any brain-relevant information.

Finally, we used partial correlations to compare vision-language models (VLMs) against their language-only counterparts (LMs). We found that neither MCSE ($\rho$ = 0.00; $p$ = 0.60) nor VisualBERT ($\rho$ = 0.00; $p$ = 0.66) exhibit statistically significant correlations with brain responses once SimCSE and BERT, respectively, are regressed out. Crucially, this indicates that VLMs did not learn any additional brain-relevant information besides that already captured by their LM counterparts.
\section{Assessing Results' Robustness}
\label{sec:additional_analyses}
RSA results revealed a consistent advantage of language-only models over the multimodal ones. This finding contrasts with the expectation---shared across a great deal of work on multimodality and language modelling---that training models on diverse data modalities, as opposed to text alone, should yield more human-like language representations. In the following, we present two analyses aimed at assessing the robustness of these findings. Given that the audio-language model CLAP did not achieve a statistically significant brain alignment, we excluded it from further analyses and focused on the remaining vision-language and language-only architectures.

\subsection{Do caption-like templates result in improved brain alignment?}

As pointed out by \citet{tan2020vokenization}, image captions are examples of \textit{grounded language}, which differs from other types of natural language along many dimensions. Since the VLMs we evaluated were trained on image-caption pairs, they may have over-fitted to the language present in captions. Therefore, it is possible that the sentence templates we used to obtain contextualised word representations from the models are somehow out-of-distribution for VLMs.

To control for this potential confound, we re-extracted word representations employing different templates, whose structure was modelled around captions (e.g., \texttt{There is an <object> in a <place>}, or \texttt{A <person> is <verb in -ing> in a <place>}). These structures were identified based on a manual inspection of captions from MS-COCO, which was part of both MCSE's and VisualBERT's training.
Given the challenges of creating caption-like templates providing a fitting context for all the word stimuli, we used different sets of templates for each sub-category of words described in Section~\ref{sec:data_and_exp_model} (e.g., \texttt{There is a <food-word> on a table in a restaurant} or \texttt{A few people gathered for a <social event-word>}). We provide the complete list of templates in Appendix~\ref{app:sentence_templates}.

\begin{figure}[]
    \centering
    \includegraphics[width=\linewidth]{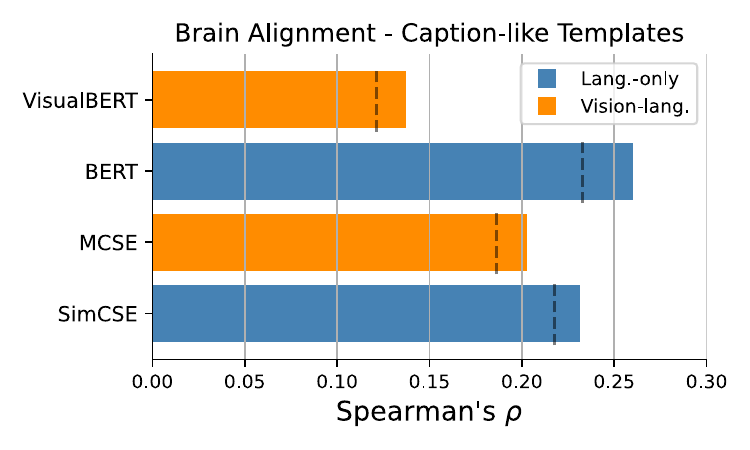}
    \caption{Spearman correlations indicating alignment between model representations extracted using caption-like templates and fMRI responses. Dotted lines indicate the initial correlations obtained with the templates from the main experiment.}
    \label{fig:rsa_brain_visual_context}
\end{figure}

The procedure for calculating brain alignment was the same as that employed in the main experiment. Spearman correlations between model-derived RDMs and the fMRI-derived RDM are displayed in Figure~\ref{fig:rsa_brain_visual_context}. All correlations are statistically significant, as well as correlation differences between models. A comparison across models confirms the trend from the main experiment: Language-only models are more brain-aligned than their vision-language counterparts. This suggests that the finding is robust and not a by-product of the templates where word stimuli were embedded.

The dotted lines in Figure~\ref{fig:rsa_brain_visual_context} allow comparing the brain alignment model representations achieve when using caption-like templates vs.~when using the templates from the main experiment. This comparison reveals that all models---not only VLMs---exhibit higher brain alignment when using caption-like templates. We interpret this as indicating that caption-like templates are not more in-distribution for VLMs, but rather provide a better-specified context that is beneficial to all models.  

\subsection{Do VLMs yield more brain-aligned representations for objects vs.~events?}

Provided that VLMs learn additional semantic information, it could be that not all word representations benefit from multimodal training to the same extent; instead, a potential advantage may be more prominent for words referring to visual contents. The words used in our main experiments include nouns from multiple semantic categories (see Section~\ref{sec:data_and_exp_model} for more details), which may largely vary in their degree of `visual-ness'. Therefore, it is possible that we did not detect additional brain-relevant information learnt by VLMs because we focused on the `wrong' words. 

To check whether this is the case, we further analysed two word subsets with different levels of concreteness. The subsets were identified by leveraging the semantic labels already present in our word set, i.e., \textit{objects} and \textit{events}.\footnote{In their supplementary materials, \citet{fernandino2022decoding} report that the average concreteness score for \textit{objects} is 4.9, while for \textit{events} it is 3.6.} We repeated RSA separately for these two word subsets following the same procedure employed in the main experiment. 

\begin{figure}[]
    \centering
    \includegraphics[width=\linewidth]{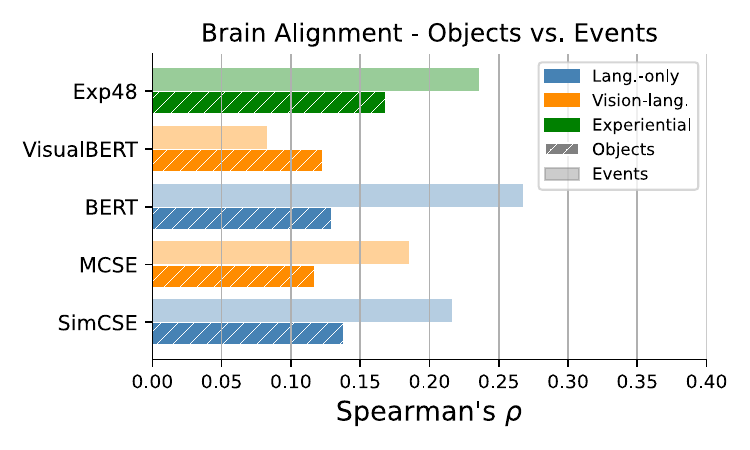}
    \caption{Spearman correlations indicating alignment between model representations and fMRI responses. Correlations are displayed separately for object-words and event-related words.}
    \label{fig:rsa_objects_events}
\end{figure}

The results of this analysis are displayed in Figure~\ref{fig:rsa_objects_events}. A first observation is that---for all models except VisualBERT---correlations are statistically significantly stronger for events than objects. This pattern was also reported by \citet{fernandino2022decoding}, who attributed it to ``higher variability of pairwise similarities for the neural representations of event concepts''. 

A second interesting result is that the model ranking we observed analysing the entire word set (BERT > SimCSE > MCSE > VisualBERT) is replicated for events but not for objects, where none of the differences between model correlations is statistically significant. While there is a negative effect overall, further training BERT on image-text pairs (as in VisualBERT) or fine-tuning it with a contrastive objective (as in SimCSE and MCSE) does not significantly alter the initial brain alignment of its object-word representations. Interestingly, \textsc{Exp48}, which we included for reference, is outperformed by BERT on events; however, it remains statistically significantly more brain-aligned than the other models on objects.

Finally, comparing vision-language models against their language-only counterparts shows that BERT and VisualBERT do not significantly differ regarding the brain alignment of their object-word representations, while SimCSE and MCSE do (with SimCSE remaining more aligned).\footnote{Note that, since we used Bonferroni corrections, this difference is statistically significant here---but not when comparing all five models---due to a change in the number of relevant comparisons (2 vs.~5).} As for event-word representations, SimCSE and BERT are, respectively, significantly more brain-aligned than MCSE and VisualBERT. These results further support the robustness of our initial finding, i.e., that LMs models are more aligned than their VL counterparts. However, the reduced gap between the two model types when considering object-word representations vs.~event-word ones suggests that VLMs \textit{do}, comparatively, learn more brain-aligned representations for objects than events.

\section{Discussion}
\label{sec:disc_concl}

While multimodal models are often expected to incorporate additional semantic aspects that language-only models cannot learn, our results reveal that their word representations are \textit{less} aligned with \textsc{Exp48} and fMRI responses than those by LMs. Moreover, within multimodal models, the vision-language ones show moderate positive correlations with \textsc{Exp48} and fMRI responses, while the audio-language one correlates weakly with \textsc{Exp48} and does not yield a significant correlation with brain responses. Below, we discuss factors that may have played a role in these partially unexpected results.  

\paragraph{Amount of training data} While being comparable in terms of learning objectives and architecture, SimCSE, MCSE and CLAP still differ in the amount of fine-tuning data. For the SimCSE--MCSE comparison, this does not appear to be a confound: Despite being fine-tuned on \textit{less} sentences than MCSE, SimCSE still proves to be \textit{more} \textsc{Exp48}- and brain-aligned. A potential reason for this may be that the grounded language employed in image captions causes a shift of semantic representations towards more concrete meanings. As for CLAP, the smaller amount of fine-tuning audio-caption pairs, together with its optimising only a CLIP-like objective (without a SimCSE-like one), may have played a role in its poor alignment. 

\paragraph{Multimodal pretraining vs. fine-tuning}  A potential explanation for the inferior performance of multimodal models could be that training on multimodal pairs is not as effective during fine-tuning as it is during pre-training. However, we verified that not even the language encoder from the powerful CLIP \cite{radford2021learning}\footnote{This model was excluded from the main experiment as it is not directly comparable with the other architectures.}---pretrained contrastively on 400M image-text pairs---yields more brain-aligned word representations than BERT and SimCSE (see Appendix~\ref{app:additional_model_results}). 

\paragraph{Models from present vs.~past studies} An interesting result was that the correlations with fMRI responses we observed for SimCSE, MCSE and BERT are higher than those achieved by the computational models (GloVe and Word2vec) tested by \citealt{fernandino2022decoding} (see Appendix~\ref{app:additional_model_results}). This finding aligns with previous work showing that transformer-based architectures are more predictive of brain responses during language processing than word-level embedding models and recurrent neural networks \cite{schrimpf2021neural}. In addition, we found that the LMs and, to a larger extent, the VLMs we tested learn brain-relevant semantic information beyond that captured by \textsc{Exp48}. This partially echoes the results by \citet{carota2024experientially}, with the difference that the computational model included in their study was strictly distributional.

\paragraph{Information captured by \textsc{Exp48}} While the ability of \textsc{Exp48} to model brain responses has been validated by previous research, it may still be a suboptimal model of perceptual experience for two main reasons.  First, all dimensions in \textsc{Exp48}, including the more perceptual ones like \textit{Colour} or \textit{Sound}, are somewhat abstract; in this sense, they may fail to capture low-level perceptual information relevant for modelling human word representations and, perhaps, learnt by multimodal models. Second, \textsc{Exp48} encodes experiential dimensions, e.g., \textit{Pleasant} or \textit{Time}, which are not strictly perceptual and may be hard, if not impossible, to learn for vision-language and audio-language models.   

\paragraph{Type of stimuli} Our study focuses on single words that are not included in longer text passages. To some extent, our results suggest that this may affect \textit{machine} language processing; indeed, we found that embedding words in sentences, as opposed to passing them to the models as is, yields more brain-aligned representations (see also Appendix~\ref{app:single_word}).
In a similar vein, the amount of context may influence \textit{human} language processing: As observed by \citet{zwaan2014embodiment}, context determines the perceptual detail of the mental simulations people engage during language comprehension. Therefore, it may be that the nouns used in the fMRI experiment did not prompt multimodal semantic knowledge enough for it to be detected in our study.

\section{Conclusions}
\label{sec:concl}

Our study provides an in-depth comparison between multimodal and language-only architectures in their ability to capture experiential semantic information and alignment with brain responses. Contrary to common assumptions, we found multimodal models to produce word representations less brain-aligned and experience-informed than language-only models. 

These results have several implications for future work. First, they invite caution against assuming that technical innovations allowing models to solve additional downstream tasks should necessarily make them more `human-like'. Second, they indicate that there is significant room for improving current computational language models so that they learn the brain-relevant experiential information they currently lack---how to concretely achieve this remains an open question.



\section*{Limitations}
\label{sec:limitations}

Our experimental setup focuses exclusively on contrastive models which are not state-of-the-art for both linguistic and multimodal downstream tasks. More recent architectures pretrained autoregressively---e.g., models from the LLaVA family \cite{liu2024improved}, Molmo \cite{deitke2024molmo}, or Qwen2.5-VL\cite{bai2025qwen2}---may exhibit different patterns. However, the complexity of their pre-training and fine-tuning steps makes it hard to set up a controlled comparison ruling out factors such as the amount of training data or training objectives. We therefore explicitly decided to not include this type of model in our investigation. This decision was further informed by 
preliminary evidence that generative vision-language models achieving stronger performance on downstream tasks are less brain-aligned than previous architectures \cite{bavaresco2024modelling}.

\section*{Acknowledgments}
\new{We thank the members of the Dialogue Modelling Group (DMG) from the University of Amsterdam and Lorenzo Proietti for the helpful feedback provided at different stages of this project.}

\new{This project was funded by the European Research Council (ERC) under the European Union’s Horizon 2020 research and innovation programme (grant agreement No.\ 819455).}

\bibliography{custom}

\appendix
\section*{Appendix}

\section{Sentence Templates}
\label{app:sentence_templates}

The  neutral sentence templates where the word stimuli were embedded in order to obtain contextualised representations from the computational models were the following:
\newline

 \noindent
 {\small 
 \texttt{Someone mentioned the <word>.} \\
 \texttt{The post was about the <word>.} \\
 \texttt{Everyone was talking about the <word>.}\\
 \texttt{They were all interested in the <word>.}\\
 \texttt{People know about the <word>.}\\
}

In one of our additional experiments (see Section 5.1), we used caption-like sentences to check whether they were more in-distribution for vision-language models and, therefore, yielded more \textsc{Exp48}- and brain-aligned representations. Below, we report the caption-like templates used for each word sub-category. \newline

\noindent Templates used for the sub-category \textit{food}: \newline

\noindent 
 {\small 
 \texttt{There is a <word> on a table in a restaurant.} \\
\texttt{A <word> is on a kitchen table.} \\
\texttt{A woman is eating a <word>.} \\
\texttt{A <word> with a few glasses around.} \\
\texttt{A close-up of a <word>.} \\
}

\noindent Templates used for the sub-category \textit{vehicle}: \newline

\noindent 
 {\small \texttt{There is one man in a <word>.} \\
\texttt{A <word> is surrounded by a few people.} \\
\texttt{A woman is posing next to a <word>.} \\
\texttt{A <word> with a young man next to it.} \\
\texttt{A close-up of a <word>.} \\
}

\noindent Templates used for the sub-category \textit{tool}: \newline

\noindent 
 {\small \texttt{There is a man holding a <word>.} \\
\texttt{A <word> is lying on the ground.} \\
\texttt{A woman is using a <word>.} \\
\texttt{A <word> with some people in the background.} \\
\texttt{A close-up of a <word>.} \\
}

\noindent Templates used for the sub-category \textit{animal}: \newline

\noindent 
 {\small 
 \texttt{There is a <word> eating voraciously.} \\
\texttt{A man is feeding a <word>.} \\
\texttt{A woman next to a <word>.} \\
\texttt{A <word> with a little girl staring at it.} \\
\texttt{A close-up of a <word>.} \\
}

\noindent Templates used for the sub-category \textit{negative event}: \newline

\noindent 
 {\small 
 \texttt{There is a crowd looking scared because of a <word>.} \\
\texttt{Many people are trying to shelter from a <word>.} \\
\texttt{A <word> happening in a big city.} \\
\texttt{A <word> with many people involved.} \\
\texttt{A picture of a <word>.} \\
}

\noindent Templates used for the sub-category \textit{social event}: \newline

\noindent 
 {\small 
 \texttt{There is a small crowd attending a <word>.} \\
\texttt{A few people are gathered for a <word>.} \\
\texttt{A <word> attended by a large group of people.} \\
\texttt{A <word> with many people involved.} \\
\texttt{A picture of a <word>.} \\
}

\noindent Templates used for the sub-category \textit{communication}: \newline

\noindent 
{\small 
\texttt{There is a small crowd at a <word>.} \\
\texttt{A few people are participating in a <word>.} \\
\texttt{A <word> in a crowded room.} \\
\texttt{A <word> with many people involved.} \\
\texttt{A picture of a <word>.} \\
}

\noindent Templates used for the sub-category \textit{sound}: \newline

\noindent 
 {\small 
 \texttt{There is a man hearing a <word>.} \\
\texttt{A few people seem to hear a <word>.} \\
\texttt{A <word> is heard by a few people.} \\
\texttt{A <word> with a few people listening to it.} \\
\texttt{A picture of a <word>.} \\
}

\section{Additional RSA Results}
\label{app:additional_results}

\begin{figure*}[h!]
    \centering
    \begin{tabular}{cc}
        \includegraphics[width=0.45\textwidth]{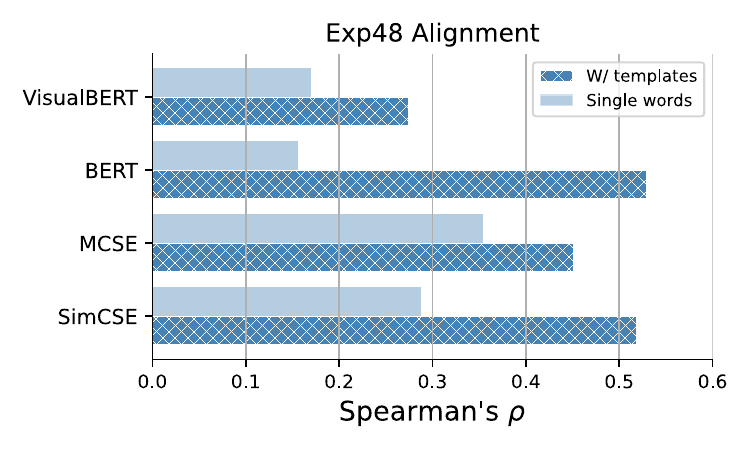} & \includegraphics[width=0.45\textwidth]{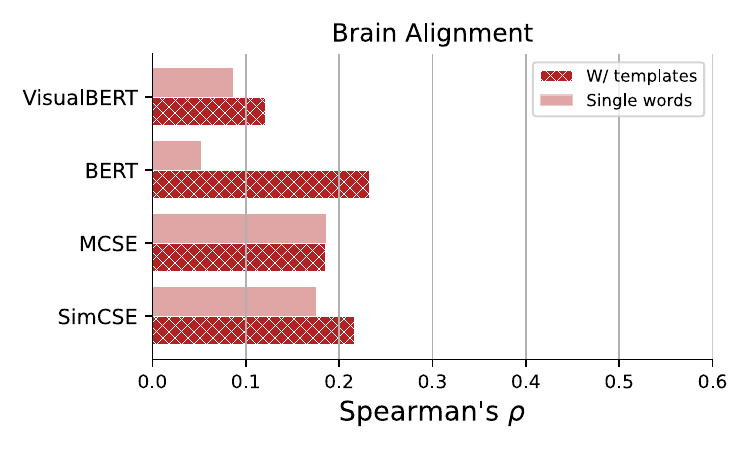} \\

    \end{tabular}
    \caption{Spearman correlations observed from model representations obtained by passing single words vs.~words embedded in templates. The left-hand panel shows the alignment with \textsc{Exp48} and the right-hand one with brain responses.}
    \label{fig:cont_non_cont_correlations}
\end{figure*}

\subsection{Single-word vs.~contextualised representations}
\label{app:single_word}

Our choice to derive word representations by including them in sentences was guided by the intuition that single words could have been an out-of-distribution input for computational models trained to output contextualised word representations. We empirically verified that representations obtained by embedding words within templates yield higher alignment than those obtained by passing single words to the models. We show the \textsc{Exp48} and brain alignment obtained with both embedding-extraction procedures in Figure~\ref{fig:cont_non_cont_correlations}.

\subsection{Layer-wise RSA results}
\label{app:layer_rsa}

\begin{figure}
    \centering
    \includegraphics[width=\linewidth]{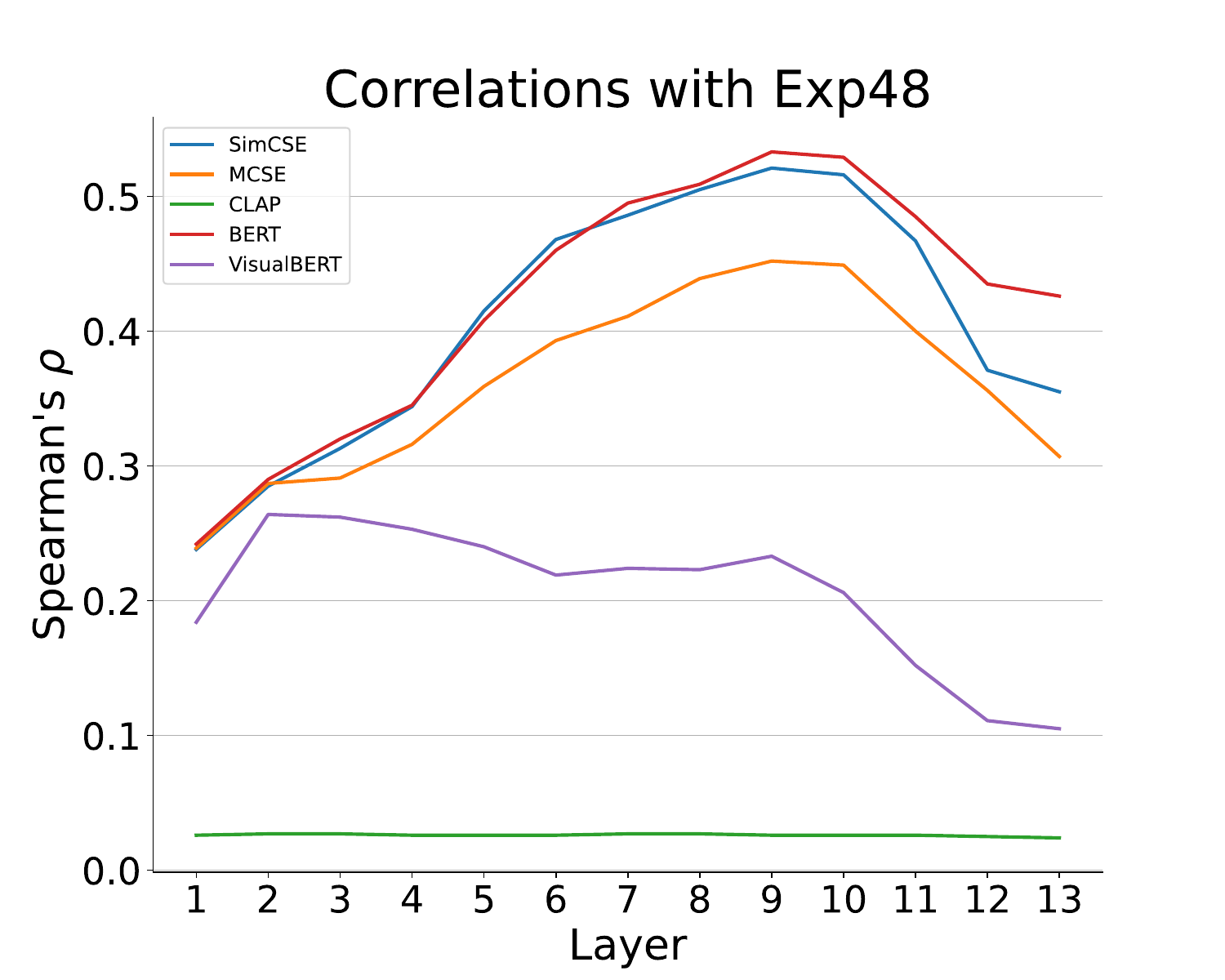}
    \caption{Spearman correlations indicating how representational similarity between model representations and \textsc{Exp48} representations changes along model layers.}
    \label{fig:rsa_exp48_layer}
\end{figure}

\begin{figure}
    \centering
    \includegraphics[width=\linewidth]{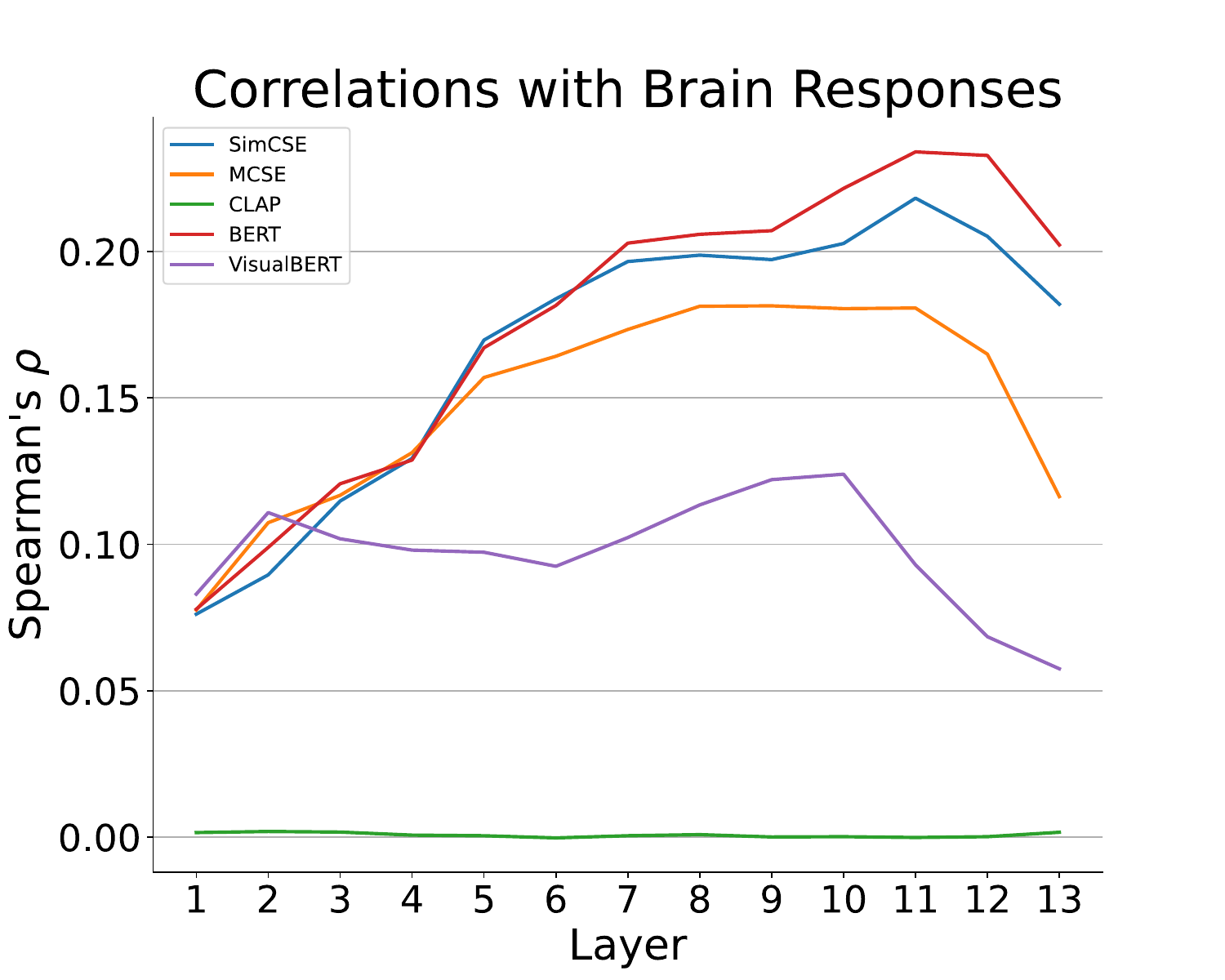}
    \caption{Spearman correlations indicating how representational similarity between model representations and brain responses changes along model layers.}
    \label{fig:rsa_brain_layer}
\end{figure}

In the main paper, we reported RSA results calculated from model representations averaged across the three layers yielding the highest alignment individually. Here, we provide a layer-wise visualisation of RSA results, which allows observing how \textsc{Exp48} vs.~ brain alignment changes throughout model layers. Specifically, layer-wise Spearman correlations against \textsc{Exp48} are displayed in Figure~\ref{fig:rsa_exp48_layer}, while those against fMRI responses are in Figure~\ref{fig:rsa_brain_layer}.

\subsection{RSA with additional baselines}
\label{app:additional_model_results}

For completeness, in Table~\ref{tab:corrs} we report RSA results including three additional models: CLIP \cite{radford2021learning}, a vision-language model pretrained contrastively on 400M image-caption pairs, and the distributional models GloVE \cite{pennington2014glove} and Word2vec \cite{word2vec}. The distributional models were originally included in \citet{fernandino2022decoding}; note that the brain correlations we report differ from the ones from \citet{fernandino2022decoding}, as they computed an average across participant-wise brain correlations, while we averaged brain RDMs across participants \textit{before} computing correlations.

\begin{table}[h]
    \centering
    \begin{tabular}{lcc} \toprule
        \textit{Model} & $\rho$ \textsc{Exp48} &  $\rho$ \textit{Brain} \\ \midrule
        SimCSE & 0.52 & 0.22\\
        MCSE & 0.45 & 0.19\\
        CLAP & 0.03 & 0.00\\
        BERT & 0.53 & 0.23\\
        VisualBERT & 0.27 & 0.12\\
        CLIP & 0.41 & 0.14\\
        GloVe & 0.45 & 0.14\\
        Word2vec & 0.42 & 0.125\\\bottomrule
    \end{tabular}
    \caption{Spearman correlations quantifying the alignment of models' representational spaces with \textsc{Exp48} and brain responses.}
    \label{tab:corrs}

\end{table}

\end{document}